\documentclass[letterpaper, 10 pt, conference, colorinlistoftodos]{ieeeconf}  % Comment this line out if you need a4paper

\IEEEoverridecommandlockouts                              % This command is only needed if 
\usepackage{graphics} % for pdf, bitmapped graphics files
\usepackage{epsfig} % for postscript graphics files
\usepackage{mathptmx} % assumes new font selection scheme installed
\usepackage{times} % assumes new font selection scheme installed
\usepackage{amsmath} % assumes amsmath package installed
\usepackage{amssymb}  % assumes amsmath package installed
\usepackage{pifont}
\usepackage{makecell} 
\usepackage[flushleft]{threeparttable}
\usepackage{booktabs}
\usepackage[backref=page]{hyperref}  
\hypersetup{
    colorlinks=true,    
    }
\usepackage{cite}
\usepackage{xcolor}

\usepackage{todonotes}
\usepackage{pdfcomment} % https://tex.stackexchange.com/questions/6306/how-to-annotate-pdf-files-generated-by-pdflatex/17004#17004

\usepackage{multirow}
\usepackage{adjustbox,lipsum}
\usepackage{tabularx}

\iftrue % Flip to true to disable comments
    \newcommand{\holger}[1]{\noindent}
    \newcommand{\shiming}[1]{\noindent}
    \newcommand{\julian}[1]{\noindent}
    \newcommand{\nan}[1]{\noindent}
    \newcommand{\red}[1]{\noindent}
    \newcommand{\blue}[1]{\noindent}
    \newcommand{\green}[1]{\noindent}
\else

    \newcommand{\holger}[1]{\pdfcomment[author=Holger,color=orange!40]{HC: #1}}
    \newcommand{\shiming}[1]{\pdfcomment[author=Shiming,color=green!40]{SW: #1}}
    \newcommand{\julian}[1]{\pdfcomment[author=Julian,color=blue!40]{JK: #1}}
    \newcommand{\nan}[1]{\pdfcomment[author=Liangliang,icon=Note,color=red!40]{Nan: #1}}
    
    \newcommand{\red}[1]{\textcolor{red}{#1}}
    \newcommand{\blue}[1]{\textcolor{blue}{#1}}
    \newcommand{\green}[1]{\textcolor{green}{#1}}
\fi

\title{\LARGE \bf
UniBEV: Multi-modal 3D Object Detection with Uniform BEV Encoders
for Robustness against Missing Sensor Modalities
}

\author{Shiming Wang, Holger Caesar, Liangliang Nan, Julian F. P. Kooij% <-this % stops a space
\thanks{All authors are with TU Delft, Delft, the Netherlands. {\tt\small \{s.wang-15, h.caesar, liangliang.nan, j.f.p.kooij\}@tudelft.nl} 
}
}

\begin{document}

\setuptodonotes{inline,color=blue!30}
\maketitle
% deleted
\thispagestyle{empty}
\pagestyle{empty}
% \thispagestyle{plain}
% \pagestyle{plain}
%%%%%%%%%%%%%%%%%%%%%%%%%%%%%%%%%%%%%%%%%%%%%%%%%%%%%%%%%%%%%%%%%%%%%%%%%%%%%%%%
\begin{abstract}
% Robustness of sensor fusion is a fundamental problem in the perception system of an autonomous driving car.
Multi-sensor object detection is an active research topic in automated driving, where the robustness of such detection models against missing sensor input (modality missing), e.g., due to a sudden sensor failure, is a critical problem that remains under studied.
%\shiming{I would like to introduce this task  explicitly in the first sentence. robustness is a too wide term as we discussed.}
In this work, we propose UniBEV, an end-to-end multi-modal 3D object detection framework designed for robustness against missing modalities:
UniBEV can operate on LiDAR plus camera input,
but also on LiDAR-only or camera-only input without retraining.
To facilitate its detector head to handle different input combinations, 
UniBEV aims to create well-aligned Bird's Eye View (BEV) feature maps from each available modality.
Unlike prior BEV-based multi-modal detection methods,
all sensor modalities follow a uniform approach to resample features from the original sensor coordinate systems to the BEV features.
%to shared BEV feature maps, \shiming{Shall we or not split the unified query thing} and (2) investigates the robustness of the fusion strategy.
% ------------- [JK] I'm HERE working
We furthermore investigate the robustness of various fusion strategies w.r.t.~missing modalities:
the commonly used feature concatenation, but also channel-wise averaging,
and a generalization to weighted averaging termed Channel Normalized Weights.
%the the average fusion over concatenation to avoid information dilution of zero-filling in the modality dropout scenario and further propose a simple yet effective fusion module, Channel Normalized Weights, to fuse aligned multi-modal BEV features.
To validate its effectiveness, we compare UniBEV to state-of-the-art BEVFusion and MetaBEV on nuScenes over all sensor input combinations.
In this setting, UniBEV achieves better performance than these baselines %\holger{remove: on average} 
for all input combinations.
%\holger{remove: significantly improving over the baselines}.
An ablation study shows the robustness benefits of fusing by weighted averaging over regular concatenation, and of sharing queries between the BEV encoders of each modality.
%UniBEV achieves $64.2 \%$ mAP with multi-modal fusion, which significantly surpasses the state-of-the-art methods, such as MetaBEV by the margin of $1.7 \%$ mAP and BEVFusion by $5.5 \%$ mAP, respectively. Notably, employing multi-modal trained weights for single-modal inference not only yields results outperforming baseline methods, especially in the camera-only scenario (supassing MetaBEV by $9.1 \%$ mAP and BEVFusion by $12.4 \%$ mAP), but also
%parallels the performance of counterpart models trained exclusively for that modality (camera: $ 35.0 \%$ vs $36.9 \%$ mAP; LiDAR: $ 58.2 \%$ vs $57.8 \%$ of mAP). This showcases UniBEV's robustness in extreme sensor failure scenarios. 
Our code is available at \url{https://github.com/tudelft-iv/UniBEV}.
\end{abstract}

% remove the brackets of L-only and C-only
% just one or two, dont need to reduce absolute numbers
% mention fusion mechanisms
% Remove the sentences md as contributions.
% We are claiming uniform approach of extraching BEV will improve robustness over existing approach.
% studying on this problem doesn't have a lot of works.
% check the typos
% don't mention cat and sum
% BEV feature encoders should be done in a unified way respect to the native coordinate sytstems of the sensor modaliteis, while the existing work relies on LSS.
% deoformable attention is a more general learning method.
%%%%%%%%%%%%%%%%%%%%%%%%%%%%%%%%%%%%%%%%%%%%%%%%%%%%%%%%%%%%%%%%%%%%%%%%%%%%%%%%
\section{INTRODUCTION}

% 1 how to represent depth
% 2 how to chose an architecture that is robust to modality dropout

\PARstart{T}{he} perception system of an intelligent vehicle typically relies on multiple sensors\cite{sun2020wod, nuscenes}, including LiDARs and cameras, to take advantage of their individual strengths and complementary nature for robust object detection.
For example, cameras provide rich texture information while LiDAR delivers dense point clouds with accurate geometric information.
Most works on multi-sensor models focus only on optimal detection performance when all sensors are available.
However, ideally a model could also be used when the input of one of its sensors is missing, i.e., modality missing, \textit{without any retraining}. 
A unified model to handle both multi-sensor and single-sensor inputs 
would facilitate gracefully degradation of its perception system in case of catastrophic sensor failure (e.g.,~broken connector) without needing to load a new set of model parameters, but also provide flexibility by supporting diverse hardware configurations (e.g.,~vehicles with different sensors).
This work therefore focuses on the design of a `robust' multi-sensor object detection model, which
in this context refers to the trained model's ability to fuse camera and LiDAR information for object detection, but also to operate on only a single modality.

\begin{figure}[t]
  \centering
  \includegraphics[width=\columnwidth]{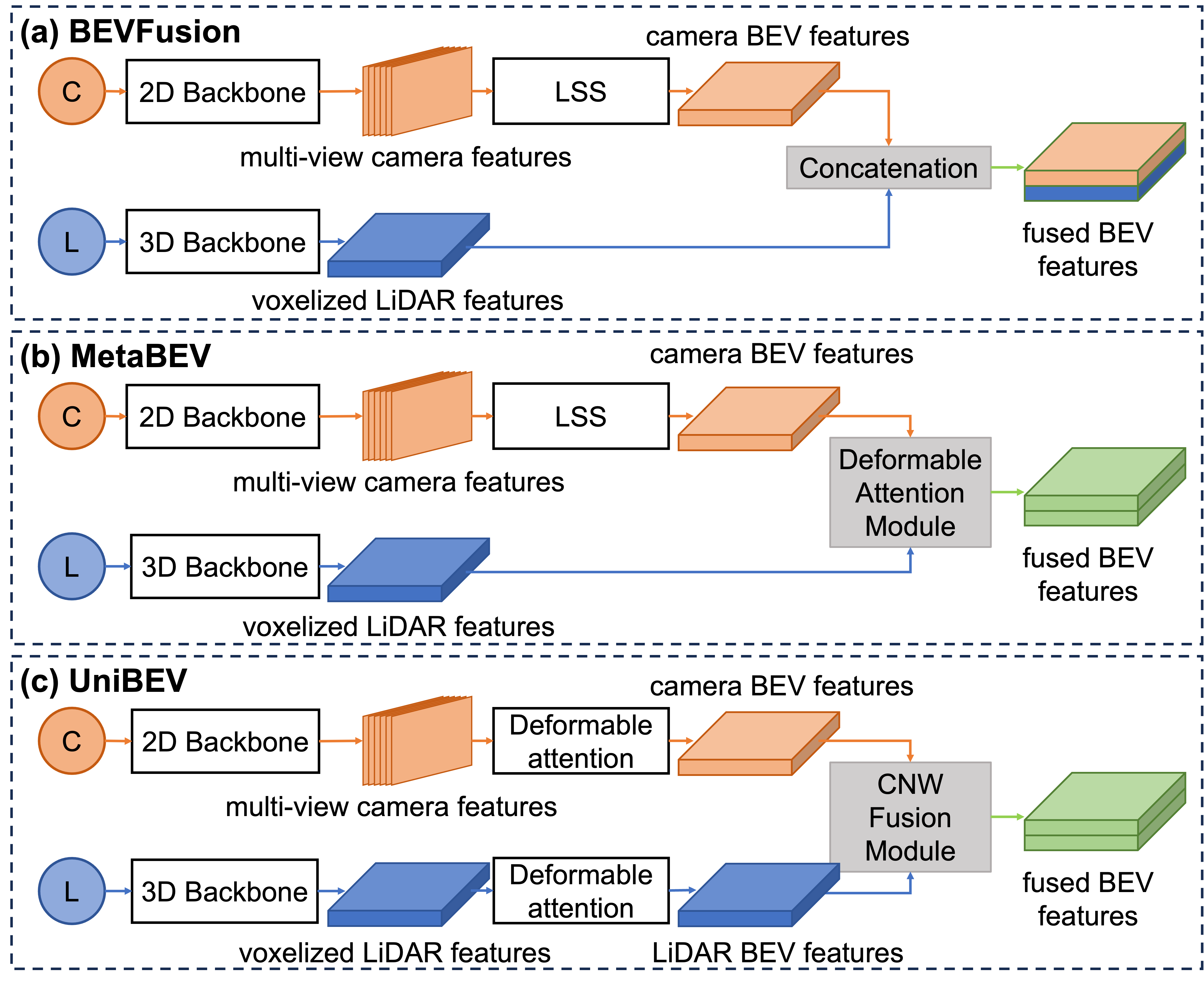}
  \caption{\textbf{Comparison of our UniBEV with other relevant works.} (a). BEVFusion \cite{liang2022bevfusion} fuses multi-modal BEV features extracted from two separate branches with concatenation. (b) MetaBEV \cite{ge2023metabev} fuses multi-modal BEV features extracted from two separate branches with a fusion module consisting of several deformable attention layers. (c) Our UniBEV extracts multi-modal BEV features from their original coordinate systems with uniform BEV encoders and fuses the BEV features with the CNW module. C and L in the figure represent the input from cameras and LiDAR. }
  \label{fig:network_comparison}
\end{figure}

% Spatial representations
Recent state-of-the-art (SotA) in multi-sensor object detection for self-driving exploits dense Bird's-Eye view (BEV) features as an intermediate representation to integrate multi-sensor information, which can then be used by a generic object detection head.
BEVFusion \cite{liu2022bevfusion} pioneers fusing multi-modal BEV features
for LiDAR and cameras.
It employs two separate branches to independently extract BEV features from each modality, and subsequently fuses these features through concatenation, as shown in Fig. \ref{fig:network_comparison} (a).
Notably, the designs of the camera and LiDAR branches in BEVFusion are non-uniform:
The camera branch relies on a (type of) Lift-Splat-Shoot (LSS) \cite{philion2020lss} component to explicitly predict the depth distribution of the image features, and map them from their camera coordinates to the spatial BEV coordinates.
In contrast, the LiDAR branch already natively expresses its features in spatial coordinates and thus does not apply additional transformations to encode its BEV features. 
This may lead to the misalignment of camera and LiDAR BEV features as they are extracted in distinct ways.
Recently MetaBEV \cite{ge2023metabev} improves over BEVFusion by replacing its concatenation with a learnable module with several deformable attention layers to better align the features, 
while still keeping BEVFusion's BEV feature encoder approach, see Fig. \ref{fig:network_comparison} (b). Since the feature misalignment is not really solved, the gain of the fusion inference is modest.
%\nan{your claim is that in previous fusion approaches the features from different modalities are not well aligned and thus the limited performance. It could be made stronger if you can quantify/visualize how the features are aligned, and compare it with that of your method. With this evidence, you can easily convince the reviewers. } 

% Concatenation is mainly adopted among state-of-the-art methods \cite{chen2022futr3d, liu2022bevfusion, liang2022bevfusion} as a fusion approach to fuse multi-modal Bird's-Eye view (BEV) or proposal features, while some other works \cite{li2022deepfusion, bai2022transfusion} applied cross-attention modules to integrate the decorative features of one modality with the features of the other main modality. Meanwhile, DeepFusion \cite{drews2022deepfusion} used a summation fusion module to achieve a simple and flexible fusion operation. However, these methods are contingent upon both modalities being present as input. The robustness of 3D object detection, particularly in scenarios with sensor absence, remains unexplored.

% Robustness

% to extract and align multi-modal features from their specific spatial coordinates
%We claim that two separate BEV encoders can result in misaligned BEV features, especially in channel dimensions.
We argue that for `robust' multi-modal 3D object detection, i.e., for both multi-sensor and single-sensor inputs without retraining, it is important that the 
BEV representations of all sensor modalities are well aligned.
We therefore propose a new end-to-end model named \textit{UniBEV}, shown in Fig. \ref{fig:network_comparison} (c),
which revisits several key architectural design choices to improve feature alignment:
%In this research,
%We therefore propose an new end-to-end model named \textbf{UniBEV}, shown in Fig. \ref{fig:network_comparison} (c).
%for robust multi-modal 3D object detection against the sensor missing failure. Our primary hypothesis posits that 
First, it uses a uniform deformable attention-based architecture for both its camera and LiDAR branch to build each sensor's BEV features,
avoiding the need for a camera-only LSS-like explicit depth prediction.
Now both branches use deformable attention to construct their BEV features,
and the learned queries can be shared between both branches to further facilitate feature alignment and provide interactions between the two branches.
Second, to fuse the multi-sensor features, we investigate using simple averaging over concatenation to avoid zeroing-out half the features when only a single sensor is available.
We also propose an extension to a learned weighted average of feature channels, called Channel Normalized Weights (CNW).
%multi-modal BEV features within a unified BEV feature space plus a simple fusion module can enhance the robustness of 3D object detectors, especially for sensor missing scenarios.
%Instead of leveraging two separate BEV encoders, UniBEV leverages a set of shared queries and unified BEV encoders to build unified BEV representations for different modalities directly from their native coordinate systems.
%Each shared query inquires about the corresponding spatial features under Bird's Eye view across different modalities with a modal-specific deformable attention-based encoder.
%Under the guidance of the geometry constraints and the unified queries, these extracted modal-specific BEV features are well aligned in both spatial and channel dimensions.
%Moreover, we thoroughly investigated the current mainstream fusion modules, such as concatenation and average, and found\nan{propose?} a generalized formulation for them. Based on this intuition, we propose a simple yet effective learnable fusion module, \textbf{C}hannel \textbf{N}ormalized \textbf{W}eights (CNW). CNW is the optimal point of the generalized formulation in comparison to concatenation and average. The general comparison between BEVFusion, MetaBEV, and our UniBEV is shown in Fig. \ref{fig:network_comparison}.

Our main contributions are as follows:

\begin{itemize}
  \item We propose UniBEV, a multi-modal 3D object detector designed for robustness against missing modalities.
  It follows a uniform approach across all modalities to encode the sensor-specific features into a shared BEV feature space to facilitate alignment between modalities.
  %This ensures robustness against missing modalities, as its detector head can work on BEV feature maps from any available modality without retraining or reloading.
  UniBEV exhibits a more robust performance than its baselines against modality missing failure without needing to load a different set of model parameters.
  \item
  When taking multi-modal data as input, UniBEV outperforms SotA multi-modal methods on the nuScenes dataset.
  For fair benchmarking, we reimplemented the closed-source MetaBEV and evaluated all multi-modal baselines under similar training conditions (e.g.~same hardware, no data augmentation).
  %We will release all source code upon paper acceptance.
  %which are obtained through uniform 
%with uniform modal-specific BEV encoders. The multi-modal BEV features are extracted from their native coordinates and well aligned in the unified BEV feature space.
  \item We investigate the impact of various feature fusion strategies: concatenation, averaging, and a simple extension to weighted averaging we call Channel Normalized Weights. For the same number of feature channels after fusion, the CNW performs better when modality missing is considered than the commonly used fusion by feature concatenation.
%to avoid zero-filling for concatenation when modality dropout is applied, and further propose a simple yet effective module, the CNW module, to better leverage complementary information across diverse modalities.
% \item We investigate the impact of shared BEV queries between the BEV encoders of all modalities, compared to using separate queries, and demonstrate consistent small improvements for all input combinations.
\end{itemize}

\begin{figure*}[t]
  \centering
  \includegraphics[width=\textwidth]{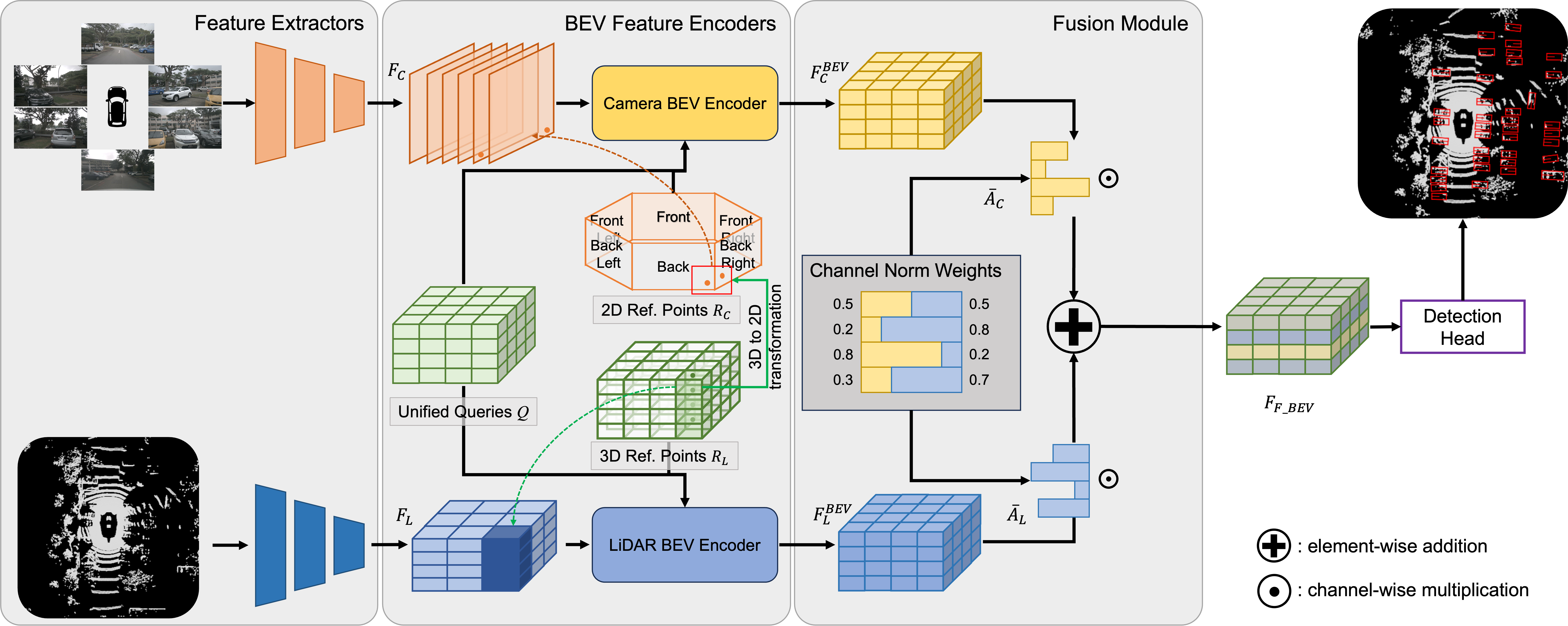}
  \caption{\textbf{The overall architecture of the UniBEV framework.} 
  1). Multi-view images and point clouds are processed through their respective backbones to generate multi-modal features. 2). A predefined set of grid-shaped BEV queries, shared across modalities, is utilized. Guided by these shared BEV queries, modality-specific BEV encoders further refine the camera and LiDAR features independently to establish aligned BEV features. These encoders are constructed using deformable attention modules and accept unified queries, relevant reference points, and the backbone-extracted features as inputs. 3). The camera and LiDAR BEV features are fused along the channels according to the learned CNW values.}
  \label{fig:unibev}
\end{figure*}
\section{RELATED WORK}

Every sensor type has specific limitations for real-world driving scenarios, hence multi-modal 3D object detection has gained great attention in recent years, especially the fusion between cameras and LiDAR. Nevertheless, the alignment between camera and LiDAR features is challenging as they are defined in different coordinates. Some methods fuse multi-modal features directly from their native coordinates with some specially designed components, such as attention \cite{vaswani2017attention, chen2022autoalign, chen2022autoalignv2, bai2022transfusion, li2022bevdepth}. DeepFusion \cite{li2022deepfusion} is the representative work among them, it simply performs cross-attention on multi-modal features with LiDAR features as queries and with camera features as keys and values. FUTR3D \cite{chen2022futr3d} brings the idea of DETR3D \cite{wang2022detr3d} and Object DGCNN \cite{obj-dgcnn} into the multi-modal domain. With the prior knowledge of camera extrinsics, FUTR3D leverages the shared object queries interacting with multi-view image features and LiDAR BEV features to sample instance-level features.

Other SotA methods build a unified intermediate BEV representation to align and fuse multi-modal features \cite{yang2022deepinteraction, hu2023ealss}.
A benefit of such BEV representations is they can serve various tasks through distinct network heads, such as simultaneous object detection and BEV map segmentation \cite{liu2022bevfusion, li2022bevformer, borse2023xalign, saha2022translating, philion2020lss}.
BEVFusion \cite{liu2022bevfusion, liang2022bevfusion} deploys Lift-Splat-Shoot (LSS) \cite{philion2020lss} to predict image depth distributions and project the image features in BEV, and deploys a regular point voxelization method, such as PointPillars \cite{lang2019pointpillars} or CenterPoint \cite{yin2021centerpoint}, to extract BEV features from LiDAR point clouds.
The multi-modal BEV feature maps are fused by concatenation.
MetaBEV \cite{ge2023metabev} upgrades the fusion module of BEVFusion to a deformable attention-based fusion block. \cite{drews2022deepfusion} uses a simple summation fusion module to integrate cross-modal BEV features and achieve convincing performance at a long distance.
%The construction of BEV features offers significant advantages for autonomous driving research, given its adaptability across various tasks using distinct heads. 
In this study, our primary emphasis is on the alignment within the BEV feature domain to address the challenges posed by missing sensor scenarios.

A few works have looked into increasing robustness against modality missing \cite{yang2018hdnet, yan2023cmt, ge2023metabev}.
These improve robustness by applying \textit{Modality Dropout} during training,
i.e.,~some portions of the training samples are presented without the input of one of the sensors.
%Since the introduction of modality dropout by HDNet \cite{yang2018hdnet}, several recent studies \cite{yan2023cmt, ge2023metabev} have adopted it to counteract sensor absences.
In this paper, we focus on this Modality Dropout setting, and explore a model's holistic test performance when presented with both or only one input modality.
 
 % Most multi-modal 3D object detection works (i.e., \cite{liu2022bevfusion, liang2022bevfusion, chen2022futr3d}) pay a lot of attention to alignment in the unified spatial feature space but rarely discuss alignment in the channel feature space. In this work, we propose to use shared BEV queries and unified BEV encoders to extract and align features in both spatial and channel-wise dimensions.

\section{METHODOLOGY}
We now describe our new architecture, UniBEV, for robust LiDAR-camera 3D object detection.
As illustrated in Fig. \ref{fig:unibev}, UniBEV consists of four parts: features extractors, uniform BEV encoders, a fusion module, and the detection head.
Each part will be described in the following subsections.
% First, the multi-view images and LiDAR point clouds are fed into two independent feature extractors to get modal-specific features in the native sensor coordinates. 
% A learnable grid-shaped BEV embedding is defined and initialized. This BEV embedding is shared across modalities and represents the feature space of the 3D space in the Bird's-Eye view. Under the guidance of the shared BEV queries, two modal-specific deformable attention-based BEV encoders sample, project, and align the diverse multi-modal features from their native coordinates into the shared BEV feature space. (\autoref{subsec:unibev}). Then, the BEV features of different modalities are fused with the proposed \textit{Channel Normalized Weights (CNW)} fusion module (\autoref{subsec:cna}). Finally, taking the fused BEV features as input, the detection head outputs the 3D predictions in the Bird's-Eye view. (\autoref{subsec: decoder & MD}). 

\newcommand{\lidar}[0]{\mathbf{L}} % lidar subscript
\newcommand{\cam}[0]{\mathbf{C}} % camera subscript

\newcommand{\numviews}[0]{V} % number of camera views
\newcommand{\numchannels}[0]{N} % number of feature dimensions
\newcommand{\depthsize}[0]{D} % number of vertical dimensions in z-directions for query

\newcommand{\Query}[0]{Q}
\newcommand{\Refs}[0]{R}
\newcommand{\BEV}[0]{BEV}
\newcommand{\Project}[0]{\mathsf{P}} % 3D-sensor projection function

\subsection{Feature Extractors}
For the initial feature extraction, UniBEV has a similar design to previous works \cite{li2022bevformer, wang2022detr3d}, relying on common image/point cloud backbones\cite{liu2022bevfusion, liang2022bevfusion, chen2022futr3d, yan2023cmt}.
Images from $\numviews$ camera views are fed into an image backbone,
such as ResNet \cite{he2016resnet}, resulting in image features $F^i_\cam \in \mathbb{R}^{H_\cam \times W_\cam \times \numchannels_\cam}$ for $1 \leq i \leq \numviews$, where $H_\cam \times W_\cam$ is the resolution of the feature map in the native image coordinates, and $\numchannels_\cam$ is the feature dimension. 
Similarly, the LiDAR scan is processed by a regular point cloud backbone, such as VoxelNet \cite{zhou2018voxelnet},
which voxelizes and extracts grid-shaped features in Bird's Eye view $F_\lidar \in \mathbb{R}^{H_\lidar \times W_\lidar \times \numchannels_\lidar}$, where $H_\lidar$, $W_\lidar$, and $\numchannels_\lidar$ are the spatial shape and feature dimensions of the features.

\subsection{Uniform BEV Feature Encoders}
\label{subsec:unibev}

After feature extraction, $F_\lidar$ and $F_\cam$ are still represented in different coordinate systems. 
$F_\lidar$ is expressed in 3D spatial coordinates similar to the target BEV space, while $F_\cam$ uses 2D image coordinates.
Existing methods generally further transfer the image features into the Bird's Eye view with LSS \cite{philion2020lss} and simply fuse the two BEV features through concatenation \cite{liu2022bevfusion, cai2023bevfusion4d}. 
We argue that the difference in network architecture between these branches may affect the alignment of the camera and LiDAR BEV features.
Furthermore, concatenating features requires zero-filling when one modality is missing.
As a result, the decoder head would operate on BEV features that are highly different depending on the available inputs, 
%depending on if camera-only, LiDAR-only, or camera+LiDAR BEV features are used.
which may impact its robustness to a missing modality.
UniBEV therefore implements a uniform design for all sensor modalities
for better aligned BEV features, as explained next.

%Additionally, we use a set of shared BEV queries to provide a weak interaction between two BEV encoders.
%The camera and LiDAR BEV encoders have uniform structures and both leverage the deformable attention mechanism \cite{zhu2020deformabledetr, xia2022defomabletr} to sample and project features from their native coordinates to the BEV.

\textbf{Queries:}
First, a set of learnable BEV query vectors~\cite{li2022bevformer} with associated 3D spatial locations is defined. 
These queries are shared by all modalities
(our ablation study will also consider separate queries per modality).
We define learnable parameters $\Query \in \mathbb{R}^{H\times W\times \numchannels}$ as BEV queries,
where $H \times W$ represent the 2D BEV spatial grid resolution in the vehicle's local spatial coordinates,
and $\numchannels$ is the number of channels in the BEV queries.
$\Refs \in \mathbb{R}^{\depthsize \times H \times W \times 4}$ contains the  corresponding spatial coordinates of BEV reference points in a 3D spatial grid $H \times W \times D$
as homogeneous coordinates $(x,y,z,1)$.
Note that $\depthsize$ reference locations are defined along the z-direction in the pillar of each 2D query location in $\Query$.
We shall use $R(z)$ to denote only the references at level $1 \leq z \leq \depthsize$.
%sampled from the Bird's Eye view, which represents the normalized coordinates of each grid in 3D space ($H \times W \times D$). Here $\depthsize = 4$ is the number of points sampled in the vertical z-direction in the pillar of each query in $\Query$.

\textbf{Projection:}
The BEV spatial locations $\Refs$ are projected to the original spatial coordinate system of each modality's feature map,
as shown in Fig. \ref{fig:unibev},
similar to FUTR3D \cite{chen2022futr3d}
\footnote{Recall FUTR3D is not a BEV-based detector, nor does it use deformable attention to sample camera features.}.
%Our camera BEV encoder follows the paradigm of BEVFormer \cite{li2022bevformer}.
%The learnable queries $\Query$ sample the BEV features of images $F_{\cam}$ at the sampling locations from the image features $F_\cam$ with the deformable attention-based image BEV encoder. To extract the feature for each grid cell of $\Query$ in Bird's Eye view from the corresponding pixel features of $F_\cam$ in perspective view,
Namely, for the feature map of each camera $i=\{1,2,...,\numviews\}$,
the 3D points $\Refs$ are projected to its 2D image-based coordinates $\Refs^i_\cam = \Project_\cam( \Refs, P^i )$ 
using the known camera extrinsics
expressed by its homogeneous projection matrix $P^i$.
Similarly, $\Refs_\lidar = \Project_\lidar( \Refs )$
projects the references to the LiDAR feature map's spatial coordinates,
for instance to scale the spatial resolution,
though in practice often $\Project_\lidar$ is an identity function.
%In most cases, $\Refs_\lidar$ and $\Refs$ can be regarded identical. \shiming{I add this sentence as the projection for LiDAR is not explicitly illustrated in the main figure.}
%\begin{align}
%    \label{eq:cam_proj}
%    \Refs^i_\cam &= P^i\cdot \Refs,
%\end{align}
%where $P^i \mathRbi{4}{4}$ is the homogeneous projection matrix for camera $i$.
%$\Refs^i_\cam\mathRbi{\depthsize}{4}$ is the 2D coordinate reference points in the image plane of one camera, and $\numviews$ is the number of cameras.

\textbf{Encoding:}
Finally, each modality's BEV feature map is constructed using 3 layers of deformable self-attention and deformable cross-attention between the BEV queries and sensor feature maps.
%to sample and encode the features in a shared BEV coordinate system.
%With the unified BEV queries $\Query$, the multi-view camera features $F_\cam$ and the 2D coordinate reference points $\Refs_\cam$,
%The camera BEV feature map $F^{\BEV}_{\cam}$ is obtained by several layer of deformable self-attention for the camera queries, and cross-attention between camera queries and camera features.
The feature map at the first layer of the camera BEV encoder, $F^{\BEV'}_{\cam}$,
is obtained by summing over all views where a reference is visible, and over all $D$ locations for each query~\cite{li2022bevformer},
\begin{align}
    F^{\BEV'}_{\cam} = \sum_{1 \leq i \leq \numviews}
    \sum_{1 \leq z \leq \depthsize} DeformAttn(\Query, \Refs^i_\cam(z), F^i_\cam),
\end{align}
where $DeformAttn$ is the deformable cross-attention defined in \cite{zhu2020deformabledetr, xia2022defomabletr}.
The output of the last layer is the final camera BEV feature map $F^{\BEV}_{\cam}$ 
passed to the fusion module.
% \todo{Need to add the illustration for Modality Embedding.}
% \begin{figure}[t]
%   \centering
%   \includegraphics[width=\columnwidth]{figures/detailed_structures_2.png}
%   \caption{The detailed illustration of BEV encoders and modality dropout strategy.
%   \holger{Lots of overlap with Fig. 2. Remove?}
%   \holger{Can we abstract away CNA here to make it easier?}
%   }
%   \label{fig:bev_encoders_n_md}
% \end{figure}

%As $F_\lidar$ and $\Query$ are both in Bird's Eye view, similarly to the image BEV encoder, 
Mirroring the cameras, the LiDAR BEV encoder performs the same operations,
with its first feature map likewise
% to sample $F^{\BEV}_{\lidar}$ from $F_\lidar$ and align the BEV features:
\begin{align}
\label{eq:lidar_proj}
    F^{\BEV'}_{\lidar} &=
    \sum_{1 \leq z \leq \depthsize}
    DeformAttn(\Query, \Refs_\lidar(z), F_\lidar).
\end{align}
Note that due to how deformable attention works,
both $F^{\BEV}_{\cam}$ and $F^{\BEV}_{\lidar}$ will retain the 
$H \times W \times \numchannels$ size of the initial $\Query$.

%Under the guidance of the unified queries $\Query$, they can maximally guarantee that multi-modal BEV features are aligned in the spatial dimension. % [JK]: don't make such hyperbolic claims that we can't prove!

\subsection{Fusion Module: Channel Normalized Weights}
\label{subsec:cna}
In the majority of multi-modal 3D object detection approaches \cite{liu2022bevfusion, liang2022bevfusion, cai2023bevfusion4d, yan2023cmt, chen2022futr3d}, \textit{concatenation} along the feature dimension is utilized as the fusion method to combine features from different modalities, to preserve a maximum amount of information. However, if one considers scenarios where a sensor input is missing, concatenation fusion
must compensate for the absent input to ensure the number of channels provided to the decoder does not change.
For instance, by filling the fused BEV features with placeholder values, typically all zeros.

We shall investigate a simple alternative,
namely fusing BEV feature maps by \textit{averaging} (or summing~\cite{drews2022deepfusion}) over all available modality feature maps.
On the one hand,
averaging risks diluting information from a more reliable sensor with that of the less reliable sensor.
On the other hand,
this fusion strategy
never needs to resort to placeholder values, and
ensures that the fused BEV feature map always has the same number of channels as each modality BEV feature map,
even if one input modality is missing.

We also
propose a generalization of average fusion which we call
\textit{Channel Normalized Weights (CNW)}.
The key idea is that the reliability of each channel in the feature map may differ from sensor to sensor,
and we should account for this when sensor measurements can be fused.
CNW therefore has as learnable parameters a $\numchannels$-dimensional vector $A_m$ for each modality $m$, which remains fixed after training.
%Each vector $A_m$ acts as a weight for the modality and remains fixed after training.
The $i$-th element $A_m(i)$ indicates the relative importance of modality $m$ for the fused result of the $i$-th channel.
Before fusion, weights $A_m$ are normalized (denoted $\overline{A}_m$) over all available sensor modalities such that they sum to 1 per channel.
Thus with two modalities, LiDAR and camera,
$\overline{A}_\cam(i), \overline{A}_\lidar(i) = softmax(A_\cam(i), A_\lidar(i))$, 
% and $\overline{A}_\lidar(i) = 1 - \overline{A}_\cam(i)$, 
s.t.

\begin{align}
    CNW(F^{\BEV}_{\cam}, F^{\BEV}_{\lidar}) = F^{\BEV}_{\cam} \odot \overline{A}_\cam + F^{\BEV}_{\lidar} \odot \overline{A}_\lidar,
\end{align}

where $\odot$ indicates channel-wise multiplication with implied broadcasting over the spatial dimensions.
In case only a single modality is available,
$softmax$ is applied to a single value per channel and normalization reduces to assigning full weights to that modality, e.g.,
$CNW(F^{\BEV}_{\cam}) = F^{\BEV}_{\cam}$.
%and $CNW(F^{\BEV}_{\lidar}) = F^{\BEV}_{\lidar}$.
%where $\overline{A}_\cam$ and $\overline{A}_\lidar$ are the normalized attention weights by a softmax function for camera and LiDAR features with dimension $N$.
%$F^{\BEV}_{\cam}$ and $F^{\BEV}_{\lidar}$ are modal-specific BEV features.

It is easy to see that CNW reduces to average fusion 
when all the learned channel weights in $\overline{A}_\cam$ and $\overline{A}_\lidar$ approach $\frac{1}{2}$.
On the other hand,
CNW can also 
reflect concatenation fusion
by allowing channels in the fused output to only take information from one modality if those channels' learned weights approach $0$ or $1$ only.
%In particular, when the first $N/2$ values of $\overline{A}_\cam$ are set to $0$ and the second half to $1$, and the values of $\overline{A}_\lidar$ are reversed, CNW performs a concatenation operation for two feature maps each of dimension $N/2$.
%
%As a generalized formulation of average and concatenation fusion, CNW is expected to learn a dataset property that indicates the feature distribution over channels. 
%Thus, CNW is scene-agnostic and fixed after training.
%Whether dynamic channel weights conditioned on the scene enhance performance is a question reserved for future investigation.
Intuitively, 
CNW adds a small number of learnable parameters to give the model more flexibility between these special cases, allowing it to optimize the relative importance of each modality for fusion,
and still allowing meaningful values for the single modal input. 
Our experimental results shall show UniBEV constructs BEV features with a similar magnitude distribution for each modality, ensuring that our CNW discerns the importance of different channels rather than a random scale function.

\subsection{Detection Head and Modality Dropout Strategy}
\label{subsec: decoder & MD}
Following previous works \cite{li2022bevformer, wang2022detr3d, carion2020detr, chen2022futr3d}, we cast the bounding box detection as a set prediction problem and adopt the decoder of BEVFormer \cite{li2022bevformer} for 3D object detection task.
To train the model for sensor missing failure, we deploy the common \textit{Modality Dropout} (MD) training strategy~\cite{yang2018hdnet, yan2023cmt, ge2023metabev}.
Thus during training we drop with a probability $p_{md}$ the BEV features of one of the modalities, either $F^{\BEV}_{\cam}$ or $F^{\BEV}_{\lidar}$. Furthermore, 
in case we do drop one of the modalities,
$p_L$ indicates the probability of keeping LiDAR, while $p_C = 1 - p_L$ is the probability of keeping the cameras.
Thus, the overall probability of keeping both sensors
is $1 - p_{md}$,
of only LiDAR is $p_{md} \cdot p_L$,
and of only cameras is $p_{md} \cdot p_C = p_{md} \cdot (1 - p_L)$.

\begin{table*}[ht]
\caption{Evaluation results on the nuScenes \textit{val} set (\textbf{best}/ \underline{second best}). Columns indicate the test input modalities: L+C = LiDAR and cameras, L = only LiDAR, and C = only cameras. 
Non-fusion models are provided for completeness,
using "-" where they do not apply.
Note: we trained all models ourselves, to ensure equal training strategies, and since \cite{ge2023metabev} has no public code.
This leads to lower performance compared to reported benchmark results\cite{liang2022bevfusion,ge2023metabev}, especially due to a lack of data augmentation.
}
\label{table:main}
\centering
\resizebox{\textwidth}{!}{
\begin{tabular} {l|c|cc|cc|cc|cc}
\toprule
\multirow{2}{*}{Method} & \multirow{2}{*}{Train} & \multicolumn{2}{c|}{L + C} & \multicolumn{2}{c|}{L} & \multicolumn{2}{c|}{C}             &\multicolumn{2}{c}{Summary Metric}\\ 
\cmidrule{3-10}               &     Modality              & NDS $\uparrow$ & mAP $\uparrow$ & NDS $\uparrow$ &mAP $\uparrow$ & NDS $\uparrow$ & mAP $\uparrow$ & NDS $\uparrow$ & mAP $\uparrow$ \\ 
\midrule
LSS \cite{philion2020lss}                    & C & - & - & - & -              & 33.0 & 28.1 & - & -\\
BEVFormer\_S \cite{li2022bevformer}          & C & - & - & - & -              & \textbf{46.2} & \textbf{40.9} & - & -\\
UniBEV\_C                                 & C & - & - & - & -                 & \underline{44.3} & \underline{36.9} & - & -\\ 
\midrule
PointPillars \cite{lang2019pointpillars}     & L & - & -                                   & 49.1          &   34.3                & - & - & - & -\\
CenterPoint \cite{yin2021centerpoint}        & L & - & -                                   & \textbf{65.4}          &   \underline{57.0} & - & - & - & -\\
UniBEV\_L                                 & L & - & -                                   & \underline{65.2}          &   \textbf{57.8}       & - & - & - & -\\
\midrule
BEVFusion \cite{liang2022bevfusion} & L + C (MD)    & 65.3 & 58.7                     & 60.6          &   49.1                & 29.6 & 22.6 & 51.8 & 43.5\\
MetaBEV \cite{ge2023metabev}& L + C (MD)   & \underline{67.5} & \underline{62.5} & \underline{65.2}      &  \underline{57.8}      & \underline{33.6} & \underline{25.9} & \underline{55.4} & \underline{48.7} \\
UniBEV (ours)                               & L + C (MD)    & \textbf{68.5} & \textbf{64.2}         & \textbf{65.3}      & \textbf{58.2}    & \textbf{42.4} & \textbf{35.0} & \textbf{58.7} & \textbf{52.5}\\
\bottomrule
\end{tabular}
}
\end{table*}

\section{EXPERIMENTS}

\subsection{Implementation Details}
\textbf{Dataset and Metrics.} We train and evaluate our approach on the nuScenes dataset \cite{nuscenes}. nuScences is a large-scale multi-modal driving dataset, which includes 6 cameras and a 32-beam LiDAR in the sensor suite.
%We follow the official metrics of nuScenes. 
As we target robustness against missing input modalities, we report the test performance on LiDAR+camera, on LiDAR-only, and on camera-only,
and report common mean Average Precision (mAP) and nuScenes detection score (NDS)~\cite{nuscenes}. 
As a \textit{summary metric} for `robustness to modality missing', we also report for both metrics the average performance over all these three possible sensor inputs.
For example, our \textit{summary mAP} is simply,
\begin{align} 
\label{eq:synthesis_metric}
    summary \ mAP = \frac{1}{3} {} (\: mAP_{L+C} {+}\: mAP_L {+}\: mAP_C).
\end{align} 

\textbf{Model.} 
We use ResNet-101 \cite{he2016resnet} with FPN \cite{lin2017fpn}
as UniBEV's camera feature extractor and VoxelNet \cite{zhou2018voxelnet} as its LiDAR feature extractor.
CNW is the default fusion approach. The grid shape of the unified query is set to $200 \times 200$ with $\numchannels = 256$. 
%for each single modality and both modalities is set to 0.25 and 0.5, respectively.

\textbf{Baselines.} 
Our first multi-sensor baseline is BEVFusion \cite{liu2022bevfusion},
and uses their implementation which includes 
a more powerful image backbone, Dual-Swin-Tiny \cite{liu2021SwinT}, 
and CenterPoint\cite{yin2021centerpoint} head.
Since BEVFusion uses concatenation for fusion,
we use zero-filling for MD.
Our second fusion baseline is the recent MetaBEV \cite{ge2023metabev},
which improves the concatenation approach of BEVFusion to a deformable attention-based fusion module.
Our MetaBEV implementation uses the same backbones and detector head as UniBEV
\footnote{The code of MetaBEV has not been released yet. We reproduced their BEV-encoder and fusion strategies based on the paper.
Their multi-task learning strategy was not implemented for a fair comparison.}.

To assess if the results on one sensor only are reasonable, we also evaluate related uni-modal baselines:
LSS \cite{philion2020lss},
BEVFormer\_S \cite{li2022bevformer},
PointPillars \cite{lang2019pointpillars},
and CenterPoint \cite{yin2021centerpoint}.
Additionally, we let UniBEV\_C denote the camera branch of UniBEV only trained with multi-view images, to compare to the  camera-only methods.
Likewise, UniBEV\_L is the LiDAR-only branch of UniBEV. 

\textbf{Training Details.}
Our model is trained in an end-to-end manner for 36 epochs. 
Due to the utilization of various tricks in baseline methods and the absence of publicly available code, accurately reproducing the performance reported in their papers proves to be challenging.
For a fair comparison, all the baselines are retrained with the same data pipeline as our model without data augmentation techniques, such as CBGS \cite{zhu2019cbgs}. 
% \footnote{The lack of data augmentation techniques and varied training devices result in the decreased performance of the baseline methods in our study, relative to the outcomes they presented in their papers.}. 
Every baseline model underwent training on four Nvidia A40 GPUs, with the entire training duration for each model spanning roughly one week.

Unless stated otherwise,
for all models the MD probability is $p_{md} = 0.5$,
and the probability for keeping each modality is identical, i.e., $p_L = p_C = 0.5$.
Therefore, in the whole training process, on average $50 \%$ of iterations are trained with multi-modal inputs, $25 \%$ with LiDAR-only inputs, and $25 \%$ with camera-only inputs.

The image and point cloud backbones are initialized with the weights of FCOS3D \cite{wang2021fcos3d} and CenterPoint\cite{yin2021centerpoint}, respectively.
Our model is implemented in the open-sourced MMDetection3D\cite{mmdet3d2020}. %For a fair comparison, all baseline methods are reproduced without any data augmentation techniques for both modalities.
% \julian{Let's add paragraph here the used hardware and training time. Mention GPU type, how many GPUs and how many days (roughly) to train each model. Needs to give impression that our training was thorough and a lot of effort.}

\subsection{Multi-Modal 3D Object Detection}
Table \ref{table:main} showcases the inference performance of the fusion-based detector on both multi-modal input, as well as single-modality input using the same trained weights.

\textbf{Multi-modal robustness.}
The summary metrics of our UniBEV ($58.7 \%$ NDS and $52.5 \%$ mAP) significantly surpass the baseline methods, indicating that UniBEV is more robust over varying input modalities.
On all input modalities, UniBEV outperforms its multi-modal baselines,
achieving $68.5 \%$ NDS and $64.2 \%$ mAP for LiDAR+camera fusion,
especially the difference in camera-only performance is notable.
%This indicates that UniBEV achieves top-tier multi-modal performance while maintaining outstanding robustness.
%Specifically, UniBEV achieves $68.5 \%$ NDS and $64.2 \%$ mAP in fusion inference, distincly outperforming BEVFusion \cite{liang2022bevfusion} by the margin of $3.2 \%$ NDS and $5.5 \%$ mAP.
%Employing the modality dropout strategy allows BEVFusion to work with single-modal input.
%When only LiDAR is available, UniBEV ($65.3 \%$ NDS and $58.2 \%$ mAP) can surpass BEVFusion ($60.6 \%$ NDS and $49.1 \%$ mAP) with a margin of $4.7 \%$ NDS and $9.1 \%$ mAP.
Despite employing a more powerful image backbone \cite{liang2022cbnet}, BEVFusion lags markedly behind UniBEV when only camera input is available.
%, recording $33.0\%$ vs $44.3\%$ NDS and $28.1\%$ vs $36.9\%$ mAP. % [JK]: these are LSS numbers, not BEVFusion.

\begin{figure}[ht]
  \centering
  \includegraphics[width=\columnwidth]{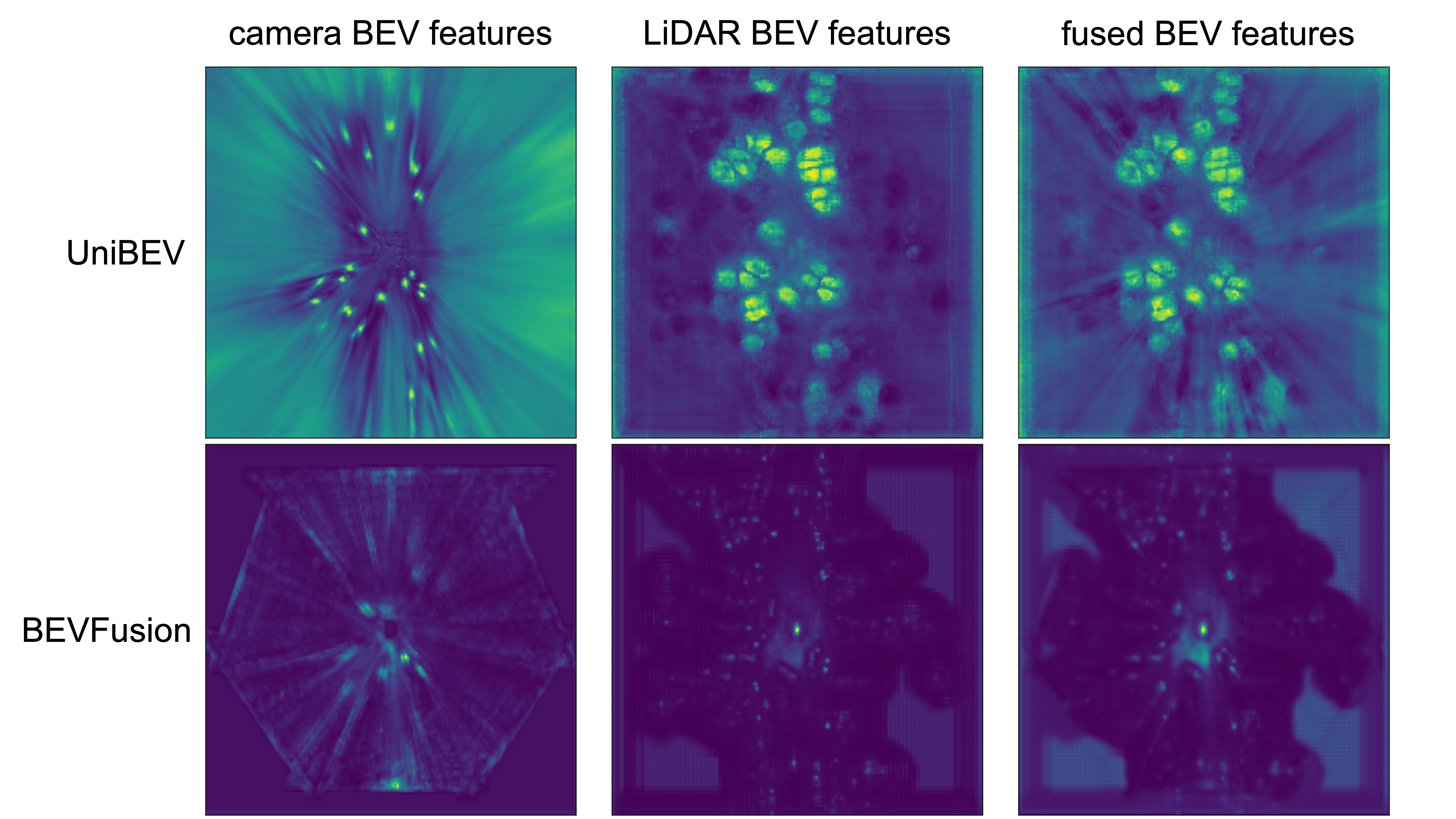}
  \caption{\textbf{Example of the BEV feature maps of different modalities
  for a single sample.}
  High intensity indicates high variance across channels at that location.
  UniBEV aligns modalities for object detection, resulting in strong responses at the same locations in both the camera and LiDAR.
  }
  \label{fig:spatial_vis}
\end{figure}

Given that the CenterPoint head of BEVFusion and our detection head exhibit comparable detection capabilities (as evidenced by the near identical performance of UniBEV\_L and CenterPoint), the performance difference for camera-only between UniBEV and BEVFusion can be attributed to the quality of BEV features and its fusion strategy. 
%
%Even if BEVFusion claims that they align both camera and LiDAR features into unified BEV representations, the camera and LiDAR BEV features are not well aligned.
Fig. \ref{fig:spatial_vis} illustrates that compared to BEVFusion, UniBEV's camera and LiDAR BEV features more clearly discern similar object locations and that these are better spatially aligned. 
%Such spatial correspondence is notably absent in BEVFusion. 
Besides, 
BEVFusion utilizes LSS \cite{philion2020lss} as the camera BEV encoder to project image features into the BEV feature space.
This enforces an inductive bias on its camera BEV features not present in its LiDAR BEV features,
as exhibited by the hexagon-shaped outline.
%While potential proposals remain visible in the LiDAR BEV features of BEVFusion, they are considerably challenging to identify within its hexagon-shaped camera BEV features.

While MetaBEV exceeds BEVFusion across all input types due to its enhanced fusion module,% (recording $67.5$ NDS and $62.5$ mAP),
~it is also outperformed overall by UniBEV.% by the gap of $1.0$ NDS and $1.7$ mAP.
~For the LiDAR-only scenario, MetaBEV does achieve comparable performance to UniBEV, which is unsurprising given the similarities in the LiDAR branch design between UniBEV and MetaBEV.
However, similarly to BEVFusion, MetaBEV also adopts LSS as its camera BEV encoder. 
Although it applies deformable attention to two BEV features
instead of our more simple CNW,
%Although MetaBEV substantially elevates its camera-only performance to $33.6 \%$ NDS and $25.9 \%$ mAP compared to BEVFusion, it still trails UniBEV by margins of $8.8 \%$ NDS and $9.1 \%$ mAP.
the BEV feature misalignment cannot be fully compensated for by merely a more powerful fusion strategy.

\begin{figure}[ht]
  \centering
  \includegraphics[width=\linewidth]{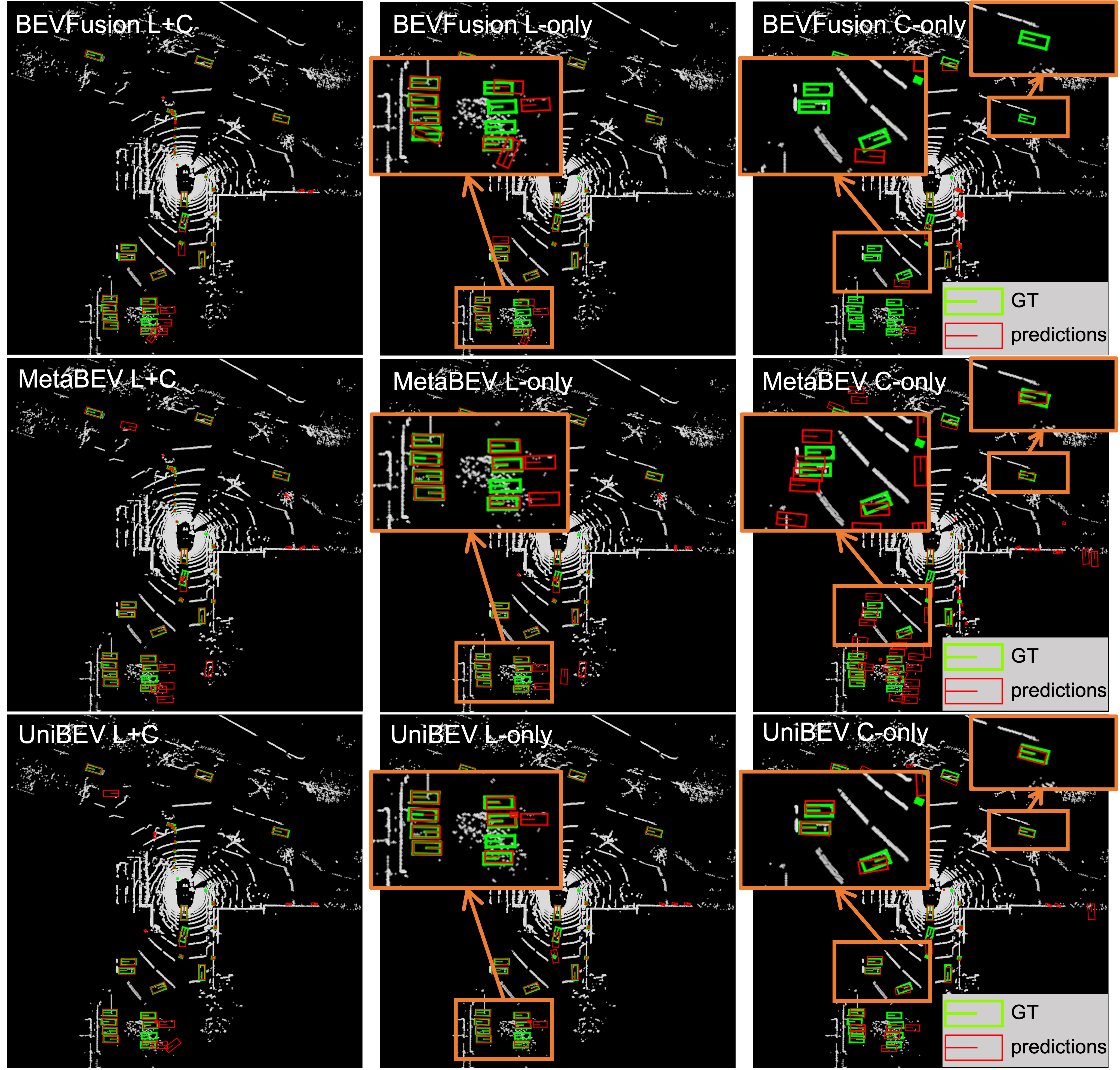}
  \caption{\textbf{Qualitative detection results on nuScenes val set.} Columns: L+C, L, and C input. Rows: BEVFusion, MetaBEV, UniBEV (ours). \textcolor{lime}{Green boxes}: ground truth; \textcolor{red}{Red boxes}: predictions. The key different zones are highlighted and zoomed in by \textcolor{orange}{orange boxes}.
}
  \label{fig:qualitative_results}
\end{figure}

\textbf{Qualitative results.}
To support the comparison between UniBEV and its multi-modal baselines,
we show some qualitative detection results in Fig. \ref{fig:qualitative_results}.
For instance, we can see that BEVFusion for camera-only suffers from various false negatives, whereas MetaBEV tends to have more false positives.
%More qualitative results are provided in the supplementary videos.

% We also demonstrate the performance of uni-modal 3D object detection frameworks. UniBEV\_C ($44.3 \%$ NDS and $36.9 \%$ mAP) can surpass LSS \cite{philion2020lss} ($33.0 \%$ NDS and $28.1 \%$ mAP) with a large margin. Due to the lack of temporal information aggregation, UniBEV\_C performs worse compared to BEVFormer\_S with NDS $46.32\%$ and mAP $40.9 \%$. Among all the LiDAR-only methods, the UniBEV\_L method ($65.2 \%$ NDS and $57.8 \%$ mAP) introduces a great performance gain compared to PointPillars \cite{lang2019pointpillars} ($49.1 \%$ NDS and $34.3 \%$ mAP) and achieves a performance comparable to CenterPoint \cite{yin2021centerpoint} ($65.4 \%$ NDS and $57.0 \%$ mAP).

\begin{figure}[ht]
  \centering
  \includegraphics[width=\columnwidth]{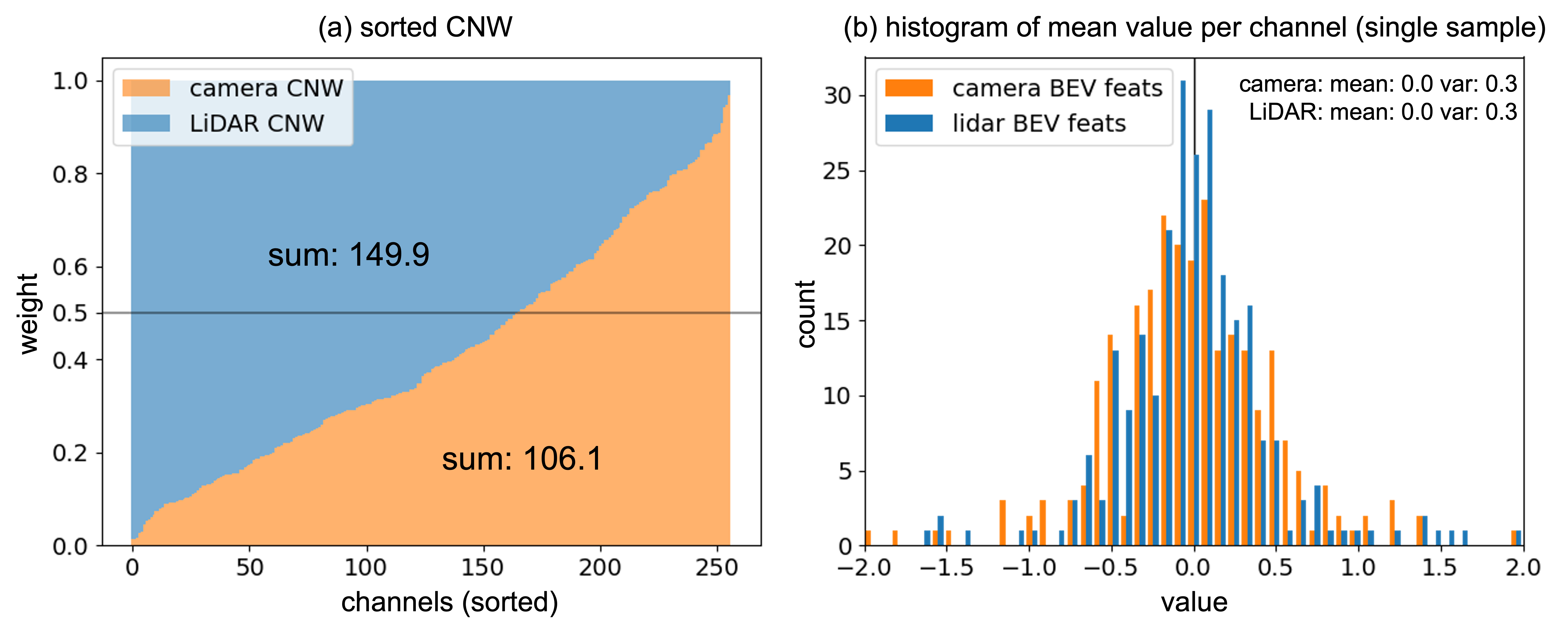}
  \caption{(a) \textbf{Visualization of CNW's learned weights, with channels sorted by weight.} (b) \textbf{Histogram for the mean value per channel of camera and LiDAR BEV features from a single sample.} The histogram demonstrates the channel-wise alignment of the two BEV feature maps. Both modalities exhibit a mean and variance of 0 and 0.3, respectively.}
  \label{fig:cna_vis}
\end{figure}

Fig. \ref{fig:cna_vis} (a) visualizes UniBEV's $\numchannels = 256$ normalized CNW weights, sorted from most LiDAR weighted to most camera weighted,
and also reports their total summed weights.
%If the model leverages the equal contributions of camera and LiDAR BEV features, the summation of CNW values for each should be 128. 
We observe the sum of camera weights is smaller than the sum of LiDAR weights ($106.1 < 149.9$). In other words, the learned fusion weights represent overall more reliance on LiDAR than on camera,
which aligns with the overall better performance of LiDAR-only models over camera-only models.
Still, we do observe quite diverse weight values.
Certainly, not all channels favor LiDAR, and few weights are close to $0.5$, 
the default for regular average fusion.
The general higher influence of LiDAR
on the fused results may also explain why the camera-only inference of UniBEV performs marginally worse than UniBEV\_C, while the LiDAR-only inference of UniBEV even slightly outperforms UniBEV\_L.

%Therefore, the camera-only performance of the fused model is sacrificed to the fusion performance.
%This observation is expected as the LiDAR information contains rich geometry information, which could contribute more to the 3D object detection task.
%Furthermore, owing to the subtle interaction facilitated by the shared BEV queries, the LiDAR-only inference performance can even marginally surpass UniBEV\_L, which is only optimized by the LiDAR input.

To validate CNW does not just scale channels to compensate for different magnitudes between LiDAR and camera BEV features,
Fig. \ref{fig:cna_vis} (b) illustrates that the  distribution of the average channel activations across the spatial map
is the same for both modalities.
%Given the well-aligned channel distributions of the camera and LiDAR BEV features, as illustrated in Fig.\ref{fig:cna_vis} (b), our proposed CNW fusion allows the model to learn the importance of distinct channels of camera and LiDAR.

% [JK] the following is better placed in the Conclusions
% -----------------------
%To sum up, thanks to the uniform BEV encoders and shared queries, UniBEV not only achieves outstanding multi-modal fusion performance but also also upholds commendable performance even in the absence of one modality. This property enhances the robustness of the perception system in an autonomous driving car.

\textbf{Inference speed.}
%The last column of Table~\ref{table:main} reports 
Finally, we measure the inference speed of all multi-modal methods using both input modalities.
Using a Nvidia V100 GPU with a batch size of 1,
we find that the average inference speed is 0.7 FPS for BEVFusion, 1.4 FPS for MetaBEV, and 1.6 FPS for our UniBEV.
%\textbf{Inference Speed.} 
%The inference speed of each model is measured on a Nvidia V100 GPU with a batch size of 1 and is expressed in frames per second (FPS). FPS metrics for multi-modal models are assessed using full input modalities. Should one modality be absent, the inference speed adjusts to that of the corresponding single-modality model.
Thus, UniBEV achieves the highest speed of all multi-modal methods by a small margin, indicating its improved performance does not come at the cost of efficiency. 
%This is particularly evident when compared to BEVFusion, which registers $65.3 \%$ NDS, $59.7 \%$ mAP and an inference speed of 0.7 FPS.
We do note that BEVFusion uses a more powerful but slower backbone~\cite{liang2022bevfusion, liang2022cbnet}.
Also, UniBEV's inference speed increases when running on multi-modal input, achieving 2.5 FPS for camera-only, and 3.9 FPS for LiDAR-only.

%While the authors of BEVFusion \cite{liang2022bevfusion} identified the LSS-based image BEV encoder as the primary latency bottleneck, MetaBEV, utilizing the same LSS-based image BEV encoder, accomplishes an inference speed of 1.4 FPS. This observation leads us to contend that the latency discrepancy is likely due to the use of different backbones \cite{liang2022cbnet}.

%In the context of single-modal models, it is observed that neither our LiDAR-only model (UniBEV\_L), nor our camera-only model (UniBEV\_C), achieves fastest inference speed compared to other single-modal models. However, it's crucial to highlight that the primary goal of this study is not to introduce state-of-the-art networks optimized for distinct modalities. Instead, our focus is on proposing the most robust model designed to operate effectively with any available modalities without retraining.

% \begin{table*}[ht]
%   \caption{Comparison of different fusion approaches on nuScenes val set. All decoder dimensions are set to 256. (\textbf{best}/ \underline{second best})}
%   \label{table:fusion_module}
%   \small
%   \centering
%   \resizebox{\textwidth}{!}{
%   \begin{tabular}{l|c|ccc|c|ccc|c}
%     \toprule
%     &   Encoder &\multicolumn{4}{c|}{mAP} &\multicolumn{4}{c}{NDS} \\
%     \cmidrule{3-10}

%     Method          & Dimensions  & L+C $\uparrow$ & L $\uparrow$ & C $\uparrow$ & Summary mAP $\uparrow$ & L+C $\uparrow$ & L $\uparrow$ & C $\uparrow$ & Summary NDS $\uparrow$ \\
%     \midrule
%     UniBEV\_cat   & 128               & 63.8          & 57.6         & 34.4         & 51.9                  & 67.8           & 64.7         &  41.7        & 58.1             \\
%     UniBEV\_avg   & 256               & \underline{64.1} &  \underline{57.6}  & \textbf{35.1} & \underline{52.3}  & \underline{68.4}    & \underline{64.9}  &  \textbf{42.6}  & \underline{58.6}             \\
%     UniBEV\_CNW   & 256               & \textbf{64.2} & \textbf{58.2} & \underline{35.0} & \textbf{52.5}  &  \textbf{68.5}    & \textbf{65.3}         &  \underline{42.4}     & \textbf{58.7}           \\
%     \bottomrule

%   \end{tabular} 
%   }
% \end{table*}

\begin{table}[ht]
  \caption{Comparison of different fusion approaches on nuScenes val set for a fixed decoder dimension of 256. The Modality Dropout strategy is applied to all models. (\textbf{best}/ \underline{second best})}
  \label{table:fusion_module}
  \small
  \centering
  \resizebox{\linewidth}{!}{
  \begin{tabular}{l|c|ccc|c}
    \toprule
    &   Encoder &\multicolumn{4}{c}{mAP}  \\
    \cmidrule{3-6}

    Method          & Dimensions  & L+C $\uparrow$ & L $\uparrow$ & C $\uparrow$ & Summary Metric $\uparrow$  \\
    \midrule
    UniBEV\_cat   & $\numchannels/2 = 128$              & 63.8          & 57.6         & 34.4         & 51.9              \\
    UniBEV\_avg   & $\numchannels = 256$               & \underline{64.1} &  \underline{57.6}  & \textbf{35.1} & \underline{52.3}     \\
    UniBEV\_CNW   & $\numchannels = 256$               & \textbf{64.2} & \textbf{58.2} & \underline{35.0} & \textbf{52.5}        \\
    \bottomrule

  \end{tabular} 
  }
\end{table}

\subsection{Ablation Study}
%To comprehensively understand the effectiveness of our proposed method,
We here discuss our ablation results for the different fusion modules, the effect of probabilities $p_L$ and $p_C$ during Modality Dropout, and the unified BEV queries.

\subsubsection{\textbf{Comparison of different fusion modules}}
We first test the performance of UniBEV with the different fusion strategies of Section~\ref{subsec:cna}: concatenation (UniBEV\_cat), average (UniBEV\_avg) and CNW (UniBEV\_CNW). 
% While the concatenation operation is prevalent in current methods, it requires zero-filling to compensate for any missing input when modality dropout is applied, which results in 
Table~\ref{table:fusion_module} demonstrates that concatenation exhibits the lowest performance with a summary mAP of $51.9 \%$.
Since a missing modality results in concatenation filling multiple fused channels with zeros,
such missing information cannot be compensated by the remaining sensor.
%We believe that this decrease in performance stems from the information dilution caused by the zero-filling process during modality dropout and the smaller dimensions of the encoder, which embeds a reduced amount of input information. 
Both UniBEV\_avg and UniBEV\_CNW avoid zero-filling in modality dropout, and subsequently elevate their performance to closely matched levels, achieving $52.3 \%$ and $52.5 \%$ summary mAP respectively.

When evaluating the performance across diverse input modalities, the L+C and L-only performances of UniBEV\_CNW improve relative to UniBEV\_avg, particularly evident in the L-only performance, but the C-only performance sees a decline.
We hypothesize that CNW effectively lets the detector head rely more on LiDAR for the final fusion result, impacting its camera-only performance.
%In contrast, average fusion mandates equal contributions from both modalities, potentially resulting in suboptimal fusion and LiDAR-only outcomes. 
Overall, the performance gap between CNW and average fusion appears minor,
and if such a trade-off is favorable for the target application remains a future research.

\subsubsection{\textbf{Effect of probabilities $p_L$ and $p_C$ during Modality Dropout}}
Next, we investigate the influence of the probabilities of keeping the different modalities during MD.
%which is determined by the probability of keeping LiDAR-only inputs, $p_L$.
Table \ref{table:md} showcases the performance of the model when changing $p_L$ and $p_C$ while keeping the probability of dropping a modality fixed to $p_{md} = 0.5$. 

The L+C performance declines significantly in the two extreme cases where $p_L = 0$ and $p_L = 1$.
However, in the other cases the L+C performance remains mostly stable, even when LiDAR-only and camera-only probabilities are unbalanced.
As expected, the performance of LiDAR-only and camera-only consistently improves as the proportion of their respective inputs used during training increases.

As shown in the table, LiDAR-only mAP increases by $12.8$ percentage points as the probability of LiDAR-only training rises from $0 \%$ to $75 \%$,
but already achieves an mAP of $45.5 \%$ even without LiDAR-only inputs during training.
Remarkably, 
training with $100 \%$ LiDAR-only during MD decreases both LiDAR-only and L+C performance compared to including $25 \%$ camera-only training iterations,
indicating that the camera-only input also regularizes the network for LiDAR.

The camera-only performance sees a substantial increase in mAP by $33$ percentage points as the proportion of camera-only training iterations increases from $0 \%$ to $100 \%$. 
However, unlike LiDAR, the camera-only performance can only reach $3.0 \%$ mAP without any training with camera-only inputs,
and does not benefit from adding LiDAR-only training iterations. 

% Particularly, as the proportion of the corresponding inputs is increased from $25 \%$ to $75 \%$, the performance improvement is more pronounced for camera-only inputs compared to LiDAR, with increases of $2.6 \%$ versus $0.5 \%$, respectively.
%This observation further supports our insights, i.e., the LiDAR information is more reliable for the 3D object detection task, allowing for exhaustive leveraging with fewer LiDAR-only training iterations.
%Furthermore, it is noteworthy that even without LiDAR-only inputs during training, the LiDAR-only inference achieves a mAP of $45.5 \%$, 
%even surpassing its camera-only performance of $36.0 \%$ mAP, despite the model being trained solely with camera-only inputs for half of the total iterations.
%while camera-only performance can only reach $3.0 \%$ mAP without any training with camera-only inputs. 
%This indicates that L-only performance can be learned during multi-modal training, but camera-only performance does not similarly benefit.

These observations further support our insights,
namely that fusion mostly relies on the more informative sensor,
in this case LiDAR,
which allows to train the LiDAR features through fusion even with few LiDAR-only training iterations.
The same is not true for the camera features,
which as the weaker modality strictly relies on MD
to make the network achieve good performance.
%This could also explain the exceptional case where increasing LiDAR-only training to $100 \%$ leads to overfitting. 
% Given the rapid convergence of the LiDAR-only detector, additional training beyond an optimal point might compromise the overall performance of the model.
This observation also suggests that especially emphasizing the weaker modality during training could enhance the overall robustness of the model, even for the other modality.
%This finding may further inspire the research into feature alignment and transfer learning across modalities.
Due to the optimal summary and L+C performance, we keep $ p_{md}=0.5$ and $p_L = p_C =0.5$ as the default for all our other experiments.
We leave studying the impact of varying the MD probability $p_{md}$ as future work.

\begin{table}[ht]
  \caption{Effect of sensor dropping probabilities on nuScenes val set. The MD probability $p_{md}$ is $0.5$. (\textbf{best}/\underline{second best}/\textbf{\textit{default setup}})}
  \label{table:md}
  \centering
  \resizebox{\linewidth}{!}{
  \begin{tabular}{ll|ccc|c}
    \toprule
    & \multicolumn{1}{l|}{}&\multicolumn{4}{c}{mAP}  \\
    \cmidrule{3-6}

    $p_L$ & $p_C$ & L+C $\uparrow$ & L $\uparrow$ & C $\uparrow$ & Summary Metric $\uparrow$  \\
    \midrule
    0 &     1 &  63.2&   45.5&  \textbf{36.0} &      48.2\\
    0.25  &  0.75   & \underline{64.0} &  57.8 &  \underline{35.8}&   \textbf{52.5}     \\
    \textbf{\textit{0.50}} & \textbf{\textit{0.50}} &  \textbf{64.2}&  \underline{58.2} &  35.0&   \textbf{52.5}      \\
    0.75  &    0.25 & 63.8 &  \textbf{58.3} &  33.2&   \underline{51.8}      \\
    1  &     0 &  60.8&  55.9&  3.0&   39.9 \\
    \bottomrule

  \end{tabular} 
  }
\end{table}

\subsubsection{\textbf{Unified Queries vs. Separate Queries}}
Finally, we compare the performance of UniBEV using shared BEV queries $\Query$ across modalities against a variant that learns separate queries for each modality.
%Both UniBEV variants adopt CNW fusion.
Table \ref{table:uni_dual_queries} 
shows the model with unified queries has a minor edge over its counterpart with separate queries across all three input combinations, as well as in the summary metric ($52.5 \%$ vs. $52.2 \%$ summary mAP).
Interestingly, 
%in the absence of the weak interaction, 
the model with separate queries 
%can, at best, 
matches the performance with UniBEV\_L (refer to Table \ref{table:main}, $57.8 \%$ mAP) while the unified queries model can slightly exceed this LiDAR-only trained counterpart.
A possible explanation is that shared queries provide
%This observation validates the effectiveness of the 
a weak interaction between the BEV encoders during training, which facilitates aligning their feature spaces. 
We conclude that the overall performance difference is only small, 
though the unified query design has the additional advantage that it requires fewer model parameters. 
For these reasons, we have adopted it as our default configuration.

\begin{table}[ht]
  \caption{Comparison between separate queries design and unified queries design on nuScenes val set. (\textbf{best}/\underline{second best})}
  \label{table:uni_dual_queries}
  \centering
  \resizebox{\linewidth}{!}{
  \begin{tabular}{l|ccc|c}
    \toprule
     &\multicolumn{4}{c}{mAP}  \\
    \cmidrule{2-5}

    Method          & L+C $\uparrow$ & L $\uparrow$ & C $\uparrow$ & Summary Metric $\uparrow$  \\
    \midrule
    separate queries      & \underline{64.0} &  \underline{57.8}  & \underline{34.9} & \underline{52.2}     \\
    unified queries       & \textbf{64.2} & \textbf{58.2} & \textbf{35.0} & \textbf{52.5}        \\
    \bottomrule

  \end{tabular} 
  }
\end{table}

\section{CONCLUSIONS}

%In this paper, 
We have presented UniBEV, a 
%fusion framework for 
multi-modal 3D object detection model
%specially 
designed with missing sensor modality inputs in mind. %against extreme sensor missing failure. 
%We hypothesize that well aligned NEB features are important for model's robustness against modality missing.
%To address misalignment between features extracted from distinct branches in existing methods, UniBEV employs uniform BEV encoders to extract and align multi-modal BEV features from their native coordinates. Meanwhile, we deploy a set of cross-modal shared queries to build an interaction between two branches.
%To avoid the zero-filling operation during modality dropout, we proposed average fusion over concatenation and a simple yet effective learnable fusion module, Channel Normalized Weights, to fuse the aligned multi-modal features.
The experiments demonstrate UniBEV's higher robustness to missing inputs
compared to SotA BEV-based detection methods, BEVFusion, and MetaBEV.
%of our proposed UniBEV through extensive experiments.
UniBEV achieves $52.5 \%$ mAP on average over all input combinations,
significantly improving over the baselines,
with BEVFusion averaging at $43.5 \%$ mAP,
and MetaBEV averaging at $48.7 \%$ mAP.
Our proposed CNW fusion approach demonstrates superior performance compared to the commonly employed concatenation. The analysis of the learned weights reveals the inherent characteristics of a fusion process, notably that the model exhibits a greater reliance on LiDAR features as compared to camera features.
% After an end-to-end multi-modal training phase, UniBEV achieves $64.2 \%$ mAP with multimodal inputs, $58.2 \%$ mAP with LiDAR-only and $35.0 \%$ mAP with camera-only, which significantly outperform the baselines.
%This performance exhibits excellent resilience against sensor missing failure, a capability that most existing methods lack. 

Future research can address dynamically adjusting channel weights, for instance based on the environmental conditions or content of the scene.
Another open question remains what properties multi-modal features should exactly possess for robustness.
%???
%devising methods to facilitate and substantiate cross-modal interaction
Finally, we will explore if UniBEV's fused BEV features also benefit other tasks, such as BEV map segmentation.
%and the interaction mechanism between different modalities.
%Our future work will focus on the further investigation of the inherent properties of multi-modal features and the interaction mechanism between different modalities.
We hope that our work will inspire further research on robust perception for autonomous driving.

\addtolength{\textheight}{0cm}   % This command serves to balance the column lengths

%%%%%%%%%%%%%%%%%%%%%%%%%%%%%%%%%%%%%%%%%%%%%%%%%%%%%%%%%%%%%%%%%%%%%%%%%%%%%%%%

%%%%%%%%%%%%%%%%%%%%%%%%%%%%%%%%%%%%%%%%%%%%%%%%%%%%%%%%%%%%%%%%%%%%%%%%%%%%%%%%

%%%%%%%%%%%%%%%%%%%%%%%%%%%%%%%%%%%%%%%%%%%%%%%%%%%%%%%%%%%%%%%%%%%%%%%%%%%%%%%%
\section*{ACKNOWLEDGMENT}
This work was supported by the 3D Urban Understanding (3DUU) Lab funded by the TU Delft AI Initiative.
%%%%%%%%%%%%%%%%%%%%%%%%%%%%%%%%%%%%%%%%%%%%%%%%%%%%%%%%%%%%%%%%%%%%%%%%%%%%%%%%

\bibliographystyle{IEEEtran}
\bibliography{IEEEabrv, reference}

\begin{thebibliography}{10}
\providecommand{\url}[1]{#1}
\csname url@rmstyle\endcsname
\providecommand{\newblock}{\relax}
\providecommand{\bibinfo}[2]{#2}
\providecommand\BIBentrySTDinterwordspacing{\spaceskip=0pt\relax}
\providecommand\BIBentryALTinterwordstretchfactor{4}
\providecommand\BIBentryALTinterwordspacing{\spaceskip=\fontdimen2\font plus
\BIBentryALTinterwordstretchfactor\fontdimen3\font minus \fontdimen4\font\relax}
\providecommand\BIBforeignlanguage[2]{{%
\expandafter\ifx\csname l@#1\endcsname\relax
\typeout{** WARNING: IEEEtran.bst: No hyphenation pattern has been}%
\typeout{** loaded for the language `#1'. Using the pattern for}%
\typeout{** the default language instead.}%
\else
\language=\csname l@#1\endcsname
\fi
#2}}

\bibitem{sun2020wod}
P.~Sun, H.~Kretzschmar, X.~Dotiwalla, A.~Chouard, V.~Patnaik, P.~Tsui, J.~Guo, Y.~Zhou, Y.~Chai, B.~Caine, \emph{et~al.}, ``Scalability in perception for autonomous driving: {W}aymo open dataset,'' in \emph{CVPR}, 2020.

\bibitem{nuscenes}
H.~Caesar, V.~Bankiti, A.~H. Lang, S.~Vora, V.~E. Liong, Q.~Xu, A.~Krishnan, Y.~Pan, G.~Baldan, and O.~Beijbom, ``{nuScenes}: A multimodal dataset for autonomous driving,'' in \emph{CVPR}, 2020.

\bibitem{liang2022bevfusion}
T.~Liang, H.~Xie, K.~Yu, Z.~Xia, Z.~Lin, Y.~Wang, T.~Tang, B.~Wang, and Z.~Tang, ``{BEVFusion}: A simple and robust lidar-camera fusion framework,'' in \emph{NeurIPS}, 2022.

\bibitem{ge2023metabev}
C.~Ge, J.~Chen, E.~Xie, Z.~Wang, L.~Hong, H.~Lu, Z.~Li, and P.~Luo, ``{MetaBEV}: Solving sensor failures for bev detection and map segmentation,'' \emph{arXiv preprint arXiv:2304.09801}, 2023.

\bibitem{liu2022bevfusion}
Z.~Liu, H.~Tang, A.~Amini, X.~Yang, H.~Mao, D.~Rus, and S.~Han, ``{BEVFusion}: Multi-task multi-sensor fusion with unified bird's-eye view representation,'' in \emph{ICRA}, 2023.

\bibitem{philion2020lss}
J.~Philion and S.~Fidler, ``Lift, splat, shoot: Encoding images from arbitrary camera rigs by implicitly unprojecting to 3d,'' in \emph{ECCV}, 2020.

\bibitem{vaswani2017attention}
A.~Vaswani, N.~Shazeer, N.~Parmar, J.~Uszkoreit, L.~Jones, A.~N. Gomez, {\L}.~Kaiser, and I.~Polosukhin, ``Attention is all you need,'' in \emph{NeurIPS}, 2017.

\bibitem{chen2022autoalign}
Z.~Chen, Z.~Li, S.~Zhang, L.~Fang, Q.~Jiang, F.~Zhao, B.~Zhou, and H.~Zhao, ``{AutoAlign}: Pixel-instance feature aggregation for multi-modal 3d object detection,'' in \emph{IJCAI}, 2022.

\bibitem{chen2022autoalignv2}
Z.~Chen, Z.~Li, S.~Zhang, L.~Fang, Q.~Jiang, and F.~Zhao, ``{AutoAlignV2}: Deformable feature aggregation for dynamic multi-modal 3d object detection,'' in \emph{ECCV}, 2022.

\bibitem{bai2022transfusion}
X.~Bai, Z.~Hu, X.~Zhu, Q.~Huang, Y.~Chen, H.~Fu, and C.-L. Tai, ``Transfusion: Robust lidar-camera fusion for 3d object detection with transformers,'' in \emph{CVPR}, 2022.

\bibitem{li2022bevdepth}
Y.~Li, Z.~Ge, G.~Yu, J.~Yang, Z.~Wang, Y.~Shi, J.~Sun, and Z.~Li, ``{BEVDepth}: Acquisition of reliable depth for multi-view 3d object detection,'' \emph{arXiv preprint arXiv:2206.10092}, 2022.

\bibitem{li2022deepfusion}
Y.~Li, A.~W. Yu, T.~Meng, B.~Caine, J.~Ngiam, D.~Peng, J.~Shen, Y.~Lu, D.~Zhou, Q.~V. Le, \emph{et~al.}, ``Deepfusion: Lidar-camera deep fusion for multi-modal 3d object detection,'' in \emph{CVPR}, 2022.

\bibitem{chen2022futr3d}
X.~Chen, T.~Zhang, Y.~Wang, Y.~Wang, and H.~Zhao, ``{Futr3d}: A unified sensor fusion framework for 3d detection,'' in \emph{CVPR}, 2022.

\bibitem{wang2022detr3d}
Y.~Wang, V.~C. Guizilini, T.~Zhang, Y.~Wang, H.~Zhao, and J.~Solomon, ``Detr3d: 3d object detection from multi-view images via 3d-to-2d queries,'' in \emph{CoRL}, 2022.

\bibitem{obj-dgcnn}
Y.~Wang and J.~M. Solomon, ``Object dgcnn: 3d object detection using dynamic graphs,'' in \emph{NeurIPS}, 2021.

\bibitem{yang2022deepinteraction}
Z.~Yang, J.~Chen, Z.~Miao, W.~Li, X.~Zhu, and L.~Zhang, ``Deepinteraction: 3d object detection via modality interaction,'' in \emph{NeurIPS}, 2022.

\bibitem{hu2023ealss}
H.~Hu, F.~Wang, J.~Su, Y.~Wang, L.~Hu, W.~Fang, J.~Xu, and Z.~Zhang, ``Ea-lss: Edge-aware lift-splat-shot framework for 3d bev object detection,'' \emph{arXiv preprint arXiv:2303.17895}, 2023.

\bibitem{li2022bevformer}
Z.~Li, W.~Wang, H.~Li, E.~Xie, C.~Sima, T.~Lu, Y.~Qiao, and J.~Dai, ``{BEVFormer}: Learning bird’s-eye-view representation from multi-camera images via spatiotemporal transformers,'' in \emph{ECCV}, 2022.

\bibitem{borse2023xalign}
S.~Borse, M.~Klingner, V.~R. Kumar, H.~Cai, A.~Almuzairee, S.~Yogamani, and F.~Porikli, ``{X-Align}: Cross-modal cross-view alignment for bird's-eye-view segmentation,'' in \emph{WCACV}, 2023.

\bibitem{saha2022translating}
A.~Saha, O.~Mendez, C.~Russell, and R.~Bowden, ``Translating images into maps,'' in \emph{ICRA}, 2022.

\bibitem{lang2019pointpillars}
A.~H. Lang, S.~Vora, H.~Caesar, L.~Zhou, J.~Yang, and O.~Beijbom, ``{PointPillars}: Fast encoders for object detection from point clouds,'' in \emph{CVPR}, 2019.

\bibitem{yin2021centerpoint}
T.~Yin, X.~Zhou, and P.~Krahenbuhl, ``Center-based 3d object detection and tracking,'' in \emph{CVPR}, 2021.

\bibitem{drews2022deepfusion}
F.~Drews, D.~Feng, F.~Faion, L.~Rosenbaum, M.~Ulrich, and C.~Gl{\"a}ser, ``Deepfusion: A robust and modular 3d object detector for lidars, cameras and radars,'' in \emph{IROS}, 2022.

\bibitem{yang2018hdnet}
B.~Yang, M.~Liang, and R.~Urtasun, ``Hdnet: Exploiting hd maps for 3d object detection,'' in \emph{Conference on Robot Learning}, 2018.

\bibitem{yan2023cmt}
J.~Yan, Y.~Liu, J.~Sun, F.~Jia, S.~Li, T.~Wang, and X.~Zhang, ``Cross modal transformer via coordinates encoding for 3d object dectection,'' \emph{arXiv preprint arXiv:2301.01283}, 2023.

\bibitem{he2016resnet}
K.~He, X.~Zhang, S.~Ren, and J.~Sun, ``Deep residual learning for image recognition,'' in \emph{CVPR}, 2016.

\bibitem{zhou2018voxelnet}
Y.~Zhou and O.~Tuzel, ``{VoxelNet}: End-to-end learning for point cloud based 3d object detection,'' in \emph{CVPR}, 2018.

\bibitem{cai2023bevfusion4d}
H.~Cai, Z.~Zhang, Z.~Zhou, Z.~Li, W.~Ding, and J.~Zhao, ``{BEVFusion4D}: Learning lidar-camera fusion under bird's-eye-view via cross-modality guidance and temporal aggregation,'' \emph{arXiv preprint arXiv:2303.17099}, 2023.

\bibitem{zhu2020deformabledetr}
X.~Zhu, W.~Su, L.~Lu, B.~Li, X.~Wang, and J.~Dai, ``Deformable {DETR}: Deformable transformers for end-to-end object detection,'' in \emph{ICLR}, 2020.

\bibitem{xia2022defomabletr}
Z.~Xia, X.~Pan, S.~Song, L.~E. Li, and G.~Huang, ``Vision transformer with deformable attention,'' in \emph{CVPR}, 2022.

\bibitem{carion2020detr}
N.~Carion, F.~Massa, G.~Synnaeve, N.~Usunier, A.~Kirillov, and S.~Zagoruyko, ``End-to-end object detection with transformers,'' in \emph{ECCV}, 2020.

\bibitem{lin2017fpn}
T.-Y. Lin, P.~Doll{\'a}r, R.~Girshick, K.~He, B.~Hariharan, and S.~Belongie, ``Feature pyramid networks for object detection,'' in \emph{CVPR}, 2017.

\bibitem{liu2021SwinT}
Z.~Liu, Y.~Lin, Y.~Cao, H.~Hu, Y.~Wei, Z.~Zhang, S.~Lin, and B.~Guo, ``Swin transformer: Hierarchical vision transformer using shifted windows,'' in \emph{ICCV}, 2021.

\bibitem{zhu2019cbgs}
B.~Zhu, Z.~Jiang, X.~Zhou, Z.~Li, and G.~Yu, ``Class-balanced grouping and sampling for point cloud 3d object detection,'' \emph{arXiv preprint arXiv:1908.09492}, 2019.

\bibitem{wang2021fcos3d}
T.~Wang, X.~Zhu, J.~Pang, and D.~Lin, ``Fcos3d: Fully convolutional one-stage monocular 3d object detection,'' in \emph{ICCV}, 2021.

\bibitem{mmdet3d2020}
M.~Contributors, ``{MMDetection3D: OpenMMLab} next-generation platform for general {3D} object detection,'' \url{https://github.com/open-mmlab/mmdetection3d}, 2020.

\bibitem{liang2022cbnet}
T.~Liang, X.~Chu, Y.~Liu, Y.~Wang, Z.~Tang, W.~Chu, J.~Chen, and H.~Ling, ``Cbnet: A composite backbone network architecture for object detection,'' \emph{IEEE T-IP}, 2022.

\end{thebibliography}

% \newpage
% \onecolumn
% \listoftodos
% \twocolumn

\end{document}